\documentclass[letterpaper, 10 pt, conference]{ieeeconf}  % Comment this line out if you need a4paper

\IEEEoverridecommandlockouts                              % This command is only needed if 
\usepackage{times}
\usepackage{epsfig} % for postscript graphics files
\usepackage{graphicx}
\usepackage{amsmath} % assumes amsmath package installed
\usepackage{amssymb}  % assumes amsmath package installed
\usepackage{multirow}
\usepackage{subfigure}
\usepackage{diagbox}
\usepackage{booktabs}
\usepackage[hang,flushmargin, symbol]{footmisc}

\usepackage{caption}
\usepackage{stfloats}
\usepackage{multirow}
\usepackage{subfigure}
\usepackage[hang,flushmargin, symbol]{footmisc}

\usepackage{booktabs}
\usepackage{latexsym}

\usepackage{algorithm}
\usepackage{algorithmic}

\makeatletter
\let\NAT@parse\undefined
\makeatother
\usepackage[colorlinks,
            breaklinks=true, 
            citecolor=green,
            bookmarks=false]{hyperref}
\hypersetup{
colorlinks=true,
linkcolor=blue
}

\title{\LARGE \bf
A Real-Time Framework for Domain-Adaptive \\Underwater Object Detection with Image Enhancement
}
\author{Junjie Wen$^{1,2}$, Jinqiang Cui$^{2\dag}$, Benyun Zhao$^{1}$, Bingxin Han$^{1}$, Xuchen Liu$^{1}$, Zhi Gao$^{3}$, Ben M. Chen$^{1}$% <-this % stops a space
\thanks{\noindent \footnotesize$^{1}$Department of Mechanical and Automation Engineering, the Chinese University of Hong Kong, Shatin, N.T., Hong Kong.}%
\thanks{\footnotesize$^{2}$Department of Mathematics and Theories, Peng Cheng Laboratory, Shenzhen, China.}
\thanks{\footnotesize$^{3}$School of Remote Sensing and Information Engineering, Wuhan University, Wuhan, China.}
\thanks{\footnotesize$^{\dag}$Corresponding author.}
}

\begin{document}

\maketitle
\thispagestyle{empty}
\pagestyle{empty}

\begin{abstract}
In recent years, significant progress has been made in the field of underwater image enhancement (UIE). However, its practical utility for high-level vision tasks, such as underwater object detection (UOD) in Autonomous Underwater Vehicles (AUVs), remains relatively unexplored. It may be attributed to several factors: (1) Existing methods typically employ UIE as a pre-processing step, which inevitably introduces considerable computational overhead and latency. (2) The process of enhancing images prior to training object detectors may not necessarily yield performance improvements. (3) The complex underwater environments can induce significant domain shifts across different scenarios, seriously deteriorating the UOD performance. To address these challenges, we introduce EnYOLO, an integrated real-time framework designed for simultaneous UIE and UOD with domain-adaptation capability. Specifically, both the UIE and UOD task heads share the same network backbone and utilize a lightweight design. Furthermore, to ensure balanced training for both tasks, we present a multi-stage training strategy aimed at consistently enhancing their performance. Additionally, we propose a novel domain-adaptation strategy to align feature embeddings originating from diverse underwater environments. Comprehensive experiments demonstrate that our framework not only achieves state-of-the-art (SOTA) performance in both UIE and UOD tasks, but also shows superior adaptability when applied to different underwater scenarios. Our efficiency analysis further highlights the substantial potential of our framework for onboard deployment.

\end{abstract}

%%%%%%%%%%%%%%%%%%%%%%%%%%%%%%%%%%%%%%%%%%%%%%%%%%%%%%%%%%%%%%%%%%%%%%%%%%%%%%%%
\section{INTRODUCTION}
The complex underwater environments pose serious challenges that notably degrade the quality of underwater images, limiting the capabilities of AUVs to execute high-level vision tasks like UOD~\cite{jesus2022underwater}\cite{katija2023autonomous}. Consequently, acquiring clear underwater images with UIE methods is commonly regarded as an essential prerequisite for vison-related underwater tasks. Nevertheless, despite the rapid advancements of UIE in recent years~\cite{akkaynak2019sea}\cite{wen2023syrea}, its practical application in underwater vision tasks, like UOD for AUVs, remains relatively unexplored. This can be ascribed to the following factors.

\begin{figure}[ht]
    \setlength{\belowcaptionskip}{-0.5cm}
    \centering
    \includegraphics[width=0.45\textwidth]{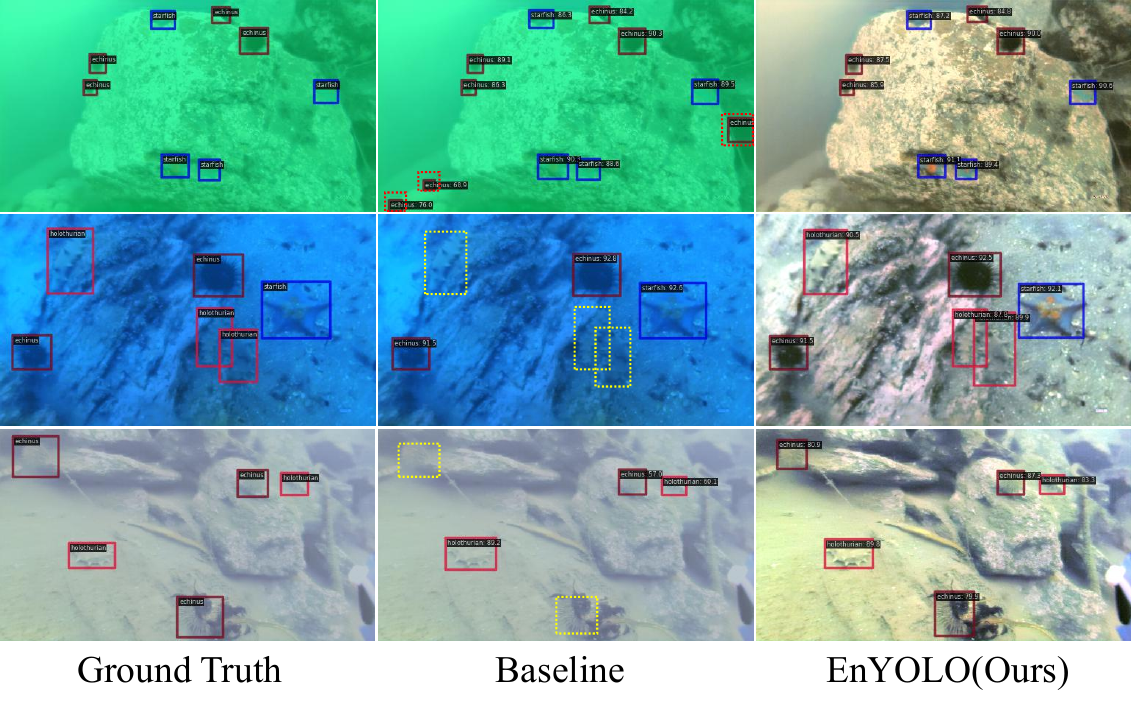}
    \caption{\small Visualization of detection results in greenish, bluish, and turbid underwater environments. Our proposed EnYOLO conducts simultaneous UIE and UOD effectively. Yellow dotted rectangles indicate missed detections, while red dotted ones represent incorrect detections. Zoom in for a better view.}
    \label{fig:demo}
\end{figure}

Firstly, using UIE as a pre-processing step for UOD unavoidably introduces additional computational latency~\cite{jiang2021underwater}\cite{Zocco2022TowardsME}, since it requires the system to await the result of UIE before executing the UOD task. Thereby, such approach significantly diminishes its suitability for real-time application. Moreover, most deep learning-based UIE techniques demand considerable computational resource, limiting their feasibility for practical integration into real-world AUV deployments.

Secondly, directly feeding the enhanced images generated by UIE to an object detector may not lead to a performance improvement. This is because most SOTA UIE techniques are primarily designed to produce visually appealing images, which may not align with the requirements of the UOD task~\cite{sun2022rethinking}. Additionally, the enhanced images may introduce artifacts that could confound the object detector~\cite{liu2022image}. 

Lastly, the complex underwater conditions frequently lead to various imaging effects (Fig.~\ref{fig:demo}), causing significant domain shift across different scenarios. It suggests that a network trained in specific underwater conditions may encounter challenges when adapting to different scenarios. While previous researchers have explored domain adaptation for UIE~\cite{wen2023syrea}\cite{chen2022domain}, there has been limited investigation into addressing the domain adaptation challenge related to UOD.

% Although some researchers have proposed frameworks aiming to jointly train UIE and UOD tasks in an end-to-end manner~\cite{fan2020dual}\cite{cheng2023joint}, such approaches often require intricate network architectures to strike a balance between the performance of both tasks, rendering them impractical for real-world application.

% Lastly, the complex underwater conditions frequently lead to various imaging effects, causing significant domain shift across different scenarios. As depicted in Fig.~\ref{fig:demo}, the captured underwater images may exhibit a greenish or bluish color cast owing to the wavelength-dependent light absorption property, and they may also show a darkened appearance as the result of insufficient illumination. It suggests that a network trained for specific underwater conditions may encounter challenges when adapting to alternative scenarios. While previous researchers have explored domain adaptation within the context of UIE~\cite{wen2023syrea}\cite{chen2022domain}, there has been limited investigation into addressing the domain adaptation challenge regarding to UOD associated with UIE.

To address the above issues, we present EnYOLO, an integrated real-time framework for simultaneously performing UIE and UOD with domain-adaptive capability. To be specific, both the UIE and UOD task heads leverage the same network backbone and employ lightweight architectures. Furthermore, we introduce a multi-stage training approach to maintain the balance in training both tasks, with the overarching goal of consistently improving their performance. Additionally, we propose a novel domain-adaptation method to mitigate the domain gaps across various underwater environments. Our main contributions are listed as follows:

\begin{itemize}
    \item We propose a unified framework capable of real-time simultaneous execution of UIE and UOD tasks.
    \item We introduce a multi-stage training strategy aimed at consistently boosting both UIE and UOD tasks.
    \item We present a novel domain-adaptation technique to mitigate the domain shift problem for UOD.
    \item Extensive experiments demonstrate the effectiveness of our framework in achieving SOTA performance for both UIE and UOD while also displaying superior adaptability across diverse underwater scenarios.
\end{itemize}

\section{RELATED WORK} \label{sec:related_work}

\subsection{Underwater Image Enhancement}
UIE techniques can be classified into traditional and learning-based approaches. While traditional methods can produce clear images by estimating backscattering and transmission under certain prior assumptions~\cite{drews2013transmission}\cite{berman2017diving}, their efficacy may decline in complex real-world scenarios. 

In contrast, learning-based approaches directly acquire the mapping from degraded underwater images to their clear counterparts, exhibiting improved adaptability in complex situations. For instance, Wang~\textit{et al.}~\cite{wang2023efficient} introduced a Swin Transformer-based~\cite{liu2021swin} UIE method that leverages local feature learning and long-range dependency modeling. Huang~\textit{et al.}~\cite{huang2023contrastive} proposed a semi-supervised UIE technique incorporating contrastive regularization to enhance the visual quality of underwater images. However, these approaches involve high computational complexity, limiting their practical feasibility for integration into real-world deployments. Although Jamieson~\textit{et al.}~\cite{jamieson2023deep} proposed a real-time algorithm that combines latest underwater image formation model with computational efficiency of deep learning frameworks, their primary focus is on visual quality enhancement, with unexplored applicability in high-level underwater vision tasks. In this study, we have designed a lightweight UIE architecture and explored its potential in the context of UOD task.

\subsection{Underwater Object Detection}
While generic object detection techniques have made remarkable advancements in diverse terrestrial applications~\cite{ren2015faster}\cite{redmon2016you}, the intricate underwater environments present substantial challenges to UOD.
% While generic object detection techniques have made remarkable advancements in diverse terrestrial applications~\cite{ren2015faster}\cite{redmon2016you}, the intricate underwater environments present substantial challenges that can detrimentally impact the effectiveness of UOD.

To enhance the performance of UOD, researchers usually leverage UIE techniques as a preliminary step to enhance image quality. For example, Jiang~\textit{et al.}~\cite{jiang2021underwater} utilized WaterNet~\cite{syariz2020waternet} to enhance underwater image quality, subsequently improving detection performance. Fan~\textit{et al.}~\cite{fan2020dual} improved detection performance by enhancing the degraded underwater images at the feature level. Enhancing the image quality before detection has also been commonly adopted in object detection for challenging terrestrial weather conditions~\cite{liu2022image}\cite{li2023detection}. However, these approaches unavoidably introduce significant computational overhead and latency. Moreover, the presence of potential artifacts in the enhanced images can lead to a drop in detection performance in some environments~\cite{sun2022rethinking}. Therefore, Cheng~\textit{et al.}~\cite{cheng2023joint} proposed a multi-task framework that jointly trains both UIE and UOD tasks in an end-to-end manner. Nevertheless, this approach relies on complex network architectures to balance the training process for both tasks, making it unfeasible for real-world deployment. In this study, we design a straightforward framework that unifies both UIE and UOD, incorporating a multi-stage training strategy aimed at consistently improving the performance of both tasks.

% Although generic object detection techniques have achieved commendable progress across various terrestrial applications~\cite{redmon2016you}\cite{ren2015faster}, the complex underwater environments could pose significant difficulties that can undermine the performance of UOD.

% \textcolor{red}{To improve the performance of UOD, researchers usually introduce UIE techniques to first improve the image quality. For example, ... However, they inevitably introduce significant computational overhead and latency. Additionally, the probable existence of artifacts in the enhanced images could potentially lead to performance drop of detection in certain environments.}

% \textcolor{red}{In this study, we propose a simple but effective multi-task framework to incorporate UIE and UOD, which can be trained in an end-to-end manner.}

\subsection{Underwater Domain Adaptation}
Domain adaptation techniques have been studied in a wide variety of tasks by mitigating the feature distribution shifts between various domains~\cite{liu2021ipmgan}\cite{li2022cross}. In the context of underwater vision tasks, domain adaptation is mainly discussed in UIE. For instance, Uplavikar~\textit{et al.}~\cite{uplavikar2019all} proposed a UIE network to handle the diversity of water types by adversarially learning the domain agnostic features. Chen~\textit{et al.} introduced a domain adaptation UIE framework by utilizing content and style separations in various domains. Wen~\textit{et al.}~\cite{wen2023syrea} developed a UIE network featuring both inter- and intra-domain adaptation strategies, enhancing its adaptability across a range of underwater scenarios. However, current techniques have primarily been confined to the realm of UIE, with limited exploration in the domain adaptation of UOD. Although Liu~\textit{et al.}~\cite{liu2020towards} proposed a domain generalization approach using WCT2 style transfer~\cite{yoo2019photorealistic} to enhance the capabilities of underwater object detector, their approach did not address the issue of domain distribution shift. In this study, we have introduced a simple but effective domain adaptation strategy for UOD, leveraging the enhanced feature embeddings derived from UIE.

% For instance, 

% \textcolor{red}{Domain adaptation has been explored widely in a variety of computer vision tasks by reducing the distribution shift between different domains. In the realm of underwater vision, domain adaptation mainly finds its application in UIE due to the variations in different underwater environments. For example, ... Despite these innovations, the existing techniques remain circuscribed to the field of UIE, leaving the domain adaptation of UOD underexplored. Although ...}

% \textcolor{red}{In this study, we introduce a simple but effective feature align method to effective reduce the domain shift between different domains. (The assumption is that the feature in enhanced images have less domain shift than the features in original degraded underwater images.)}

\begin{figure*}[th]
    \centering
    \includegraphics[width=0.92\textwidth]{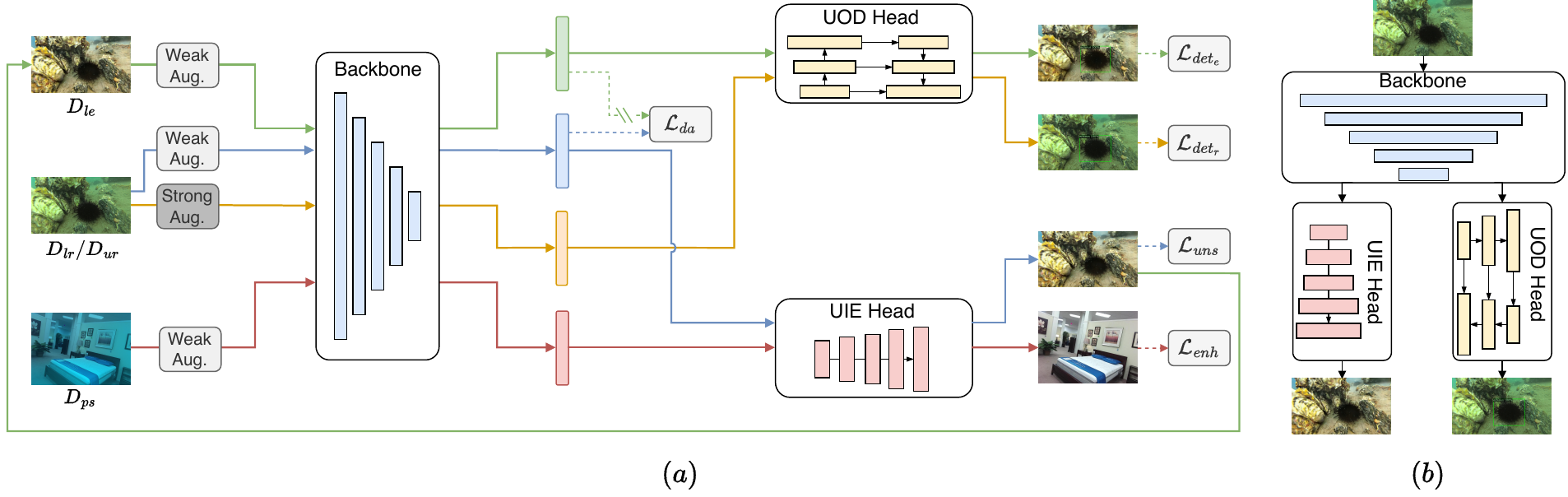}
    \captionsetup{font={small}}
    \caption{\small Overview of our proposed EnYOLO framework. (a) The schematic illustration of training process. (b) The inference process.}
    
    \label{fig:overall_arch}
\end{figure*}

\section{Proposed Method} \label{sec:method}

\subsection{Problem Definition}
Our proposed framework aims to effectively enable both UIE and UOD simultaneously. The datasets for training in this study are defined as follows. As depicted in Fig.~\ref{fig:overall_arch}\textcolor{blue}{(a)}, we use a paired synthetic underwater dataset $D_{ps}=\{(x_s, \hat{x}_s)_i, i\in[1, n_s]\}$ to facilitate the training of the UIE task; where $n_s$ represents the synthetic dataset size, $x_s$ denotes the degraded synthetic underwater image, and $\hat{x}_s$ is the corresponding clear image. For the training of UOD task, we utilize a labeled real-world underwater dataset $D_{lr}=\{(x_r, b_r, c_r)_i, i\in[1, n_r]\}$; where $n_r$ refers to the real-world dataset size, $x_r$ represents the real-world underwater image, $b_r$ indicates the bounding box annotations, and $c_r$ is the class labels. Furthermore, real-world underwater images from $D_{lr}$ also constitute an unpaired real-world underwater dataset $D_{ur}=\{(x_r)_i,i\in[1,n_r]\}$ to enhance the performance of UIE module in real-world scenarios. Additionally, the enhanced result $\tilde{x}_r$ of UIE for each $x_r$, together with their corresponding $b_r$ and $c_r$, formulate a labeled enhanced real-world dataset $D_{le}=\{(\tilde{x_r}, b_r, c_r)_i, i\in[1,n_r]\}$, which is also utilized for training the UOD task. During the inference, the network takes real-world underwater images $x_r$ and subsequently predicts both enhanced images $\tilde{x}_r$ and detection results $(\tilde{b}_r, \tilde{c}_r)$, as shown in Fig.~\ref{fig:overall_arch}\textcolor{blue}{(b)}.

% \textcolor{red}{As shown in Figure \ref{fig2}, we incorporate both synthetic and real-world underwater images to train the whole framework. We use a paired synthetic underwater images $D_{ps}=\{(x_s^i, \hat{x}_s^i), i\in[1,n_s] \}$ to train the UIE task; where $n_s$ is the quantity of the dataset, $x_s^i$ denotes the degraded synthetic underwater images in the dataset, and $\hat{x}_s^i$ is the clear references. We also use labeled real-world underwater images $D_{lr}=\{(x_r^i, b_r^i, c_r^i), i\in[1, n_r]\}$ for UOD task; where $n_r$ is the degraded real-world underwater images, $b_r^i$ indicates the bounding box annotations, and $c_r$ refers to the corresponding class labels. In addition, we also use unpaired real underwater images $D_{ur}=\{(x_r^i), i\in[1, n_r]\}$ for the UIE task. } 

\subsection{Network Overview}
As illustrated in Fig.~\ref{fig:overall_arch}, our network architecture follows a straightforward multi-task structure, comprising three components: a network backbone, a UOD head, and a UIE head. The UIE and UOD heads share the same network backbone, with no feature exchange between them during inference. Although such design is common in various multi-task network paradigms~\cite{He2017ICCV}\cite{wang2019end}, it holds particular significance for real-world underwater vision tasks. Firstly, during real-world testing, there is no dependency of the UOD head on the enhanced results from the UIE task, significantly reducing computational latency. Secondly, the decoupled nature of UOD and UIE heads introduces greater flexibility for real-world testing (Sec.~\ref{sec:exp_eff}).

\begin{figure}[h]
    \centering
    \includegraphics[width=0.45\textwidth]{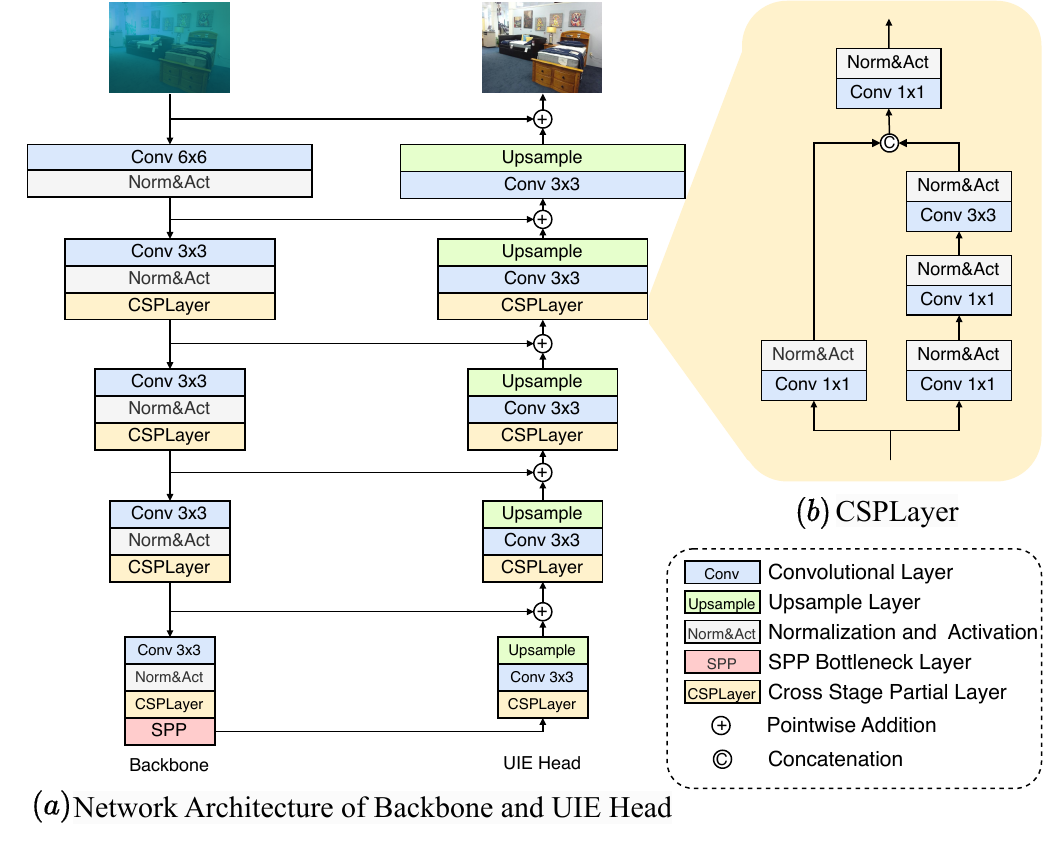}
    \caption{Network Architecture for UIE task.}
    \label{fig:uie_head}
    \vspace{-1.0em}
\end{figure}

It is clear that the network backbone and the UOD head constitute a general object detector. In order to enable real-time detection and maintain a lightweight architecture, we employ the classic one-stage object detector YOLOv5~\cite{Jocher2020YOLOv5}, which incorporates CSPDarkNet53~\cite{wang2020cspnet} as the backbone. For the UIE head, we adopt the core concept of CSPLayer~\cite{wang2020cspnet} to achieve a balance between network performance and computational efficiency. As illustrated in Fig.~\ref{fig:uie_head}, the majority of convolutional filters within the CSPLayer employ a $1\times 1$ kernel size, significantly reducing parameter count and computational demands. Moreover, the upsampling layer in our UIE head utilizes bilinear interpolation, which requires no additional network parameters and maintains computational efficiency.

\subsection{Multi-Stage Training Strategy}
The training process for both UOD and UIE is non-trivial since UIE primarily aims to enhance visual quality and fine details, while UOD mainly focuses on extracting relevant object features and their localizations. To address this issue, we introduce a multi-stage training strategy (Alg.~\ref{algo}) consisting of three training stages: the \textbf{Burn-In} stage, the \textbf{Mutual-Learning} stage, and the \textbf{Domain-Adaptation} stage.

% Although the overall architecture of our proposed network is straightforward, the training process for both UOD and UIE remains non-trivial. UIE primarily aims to enhance the visual quality and refine the detailed representations of underwater images, whereas UOD mainly focus on extracting relevant object features and localizing their positions. Consequently, the main objectives of UOD and UIE may not align with each other, making it challenging to attain a balance in the end-to-end training of both tasks.

(1) \textbf{Burn-In} Stage: In this stage, the entire network is trained to acquire fundamental capabilities in both UIE and UOD tasks. As shown in Fig.~\ref{fig:overall_arch}\textcolor{blue}{(a)}, the network is provided with paired synthetic underwater images $D_{ps}$ with weak augmentations, including random horizontal flipping and cropping, for the UIE task. Similarly, the UOD task is optimized using the labeled real-world underwater dataset $D_{lr}$ with strong augmentations, including mosaic patching, random color jittering, blurring, as well as random horizontal flipping and cropping. Since $D_{ps}$ includes reference images, and $D_{lr}$ is well-annotated, both tasks are trained in a supervised manner. During this stage, while the network acquires capability in enhancing synthetic underwater images, it may encounter difficulties when handling real-world underwater images. Similarly, although UOD is initialized with the capability to detect real-world underwater objects, it may face challenges in detecting objects across different underwater environments.

% In this stage, the whole network is trained to acquire the fundamental capabilities in both UIE and UOD tasks. As shown in Fig.~\ref{fig:overall_arch}\textcolor{blue}{(a)}, the network is fed with paired synthetic underwater images $D_{ps}$ with weak augmentations (including random horizontal flipping and cropping) for the UIE task. Likewise, the UOD task is optimized by utilizing the labeled real-world underwater dataset $D_{lr}$ with strong augmentations, encompassing mosaic patching, random color jittering, bluring, as well as random horizontal flipping and cropping. Since $D_{ps}$ includes reference images and $D_{lr}$ is well annotated, both tasks are trained in a supervised manner. Through this stage, while the network acquires capabilities in enhancing synthetic underwater images, it may encounter difficulties when handling real-world underwater images. Similarly, although UOD is initialized with the capability to detect real-world underwater objects, it may face challenges in detecting objects across different underwater environments.

(2) \textbf{Mutual-Learning} Stage: During this stage, the performance of both UIE and UOD task is further improved through mutual knowledge learning. In the case of the UIE task, knowledge from real-world underwater images is leveraged by training with unpaired real-world underwater images $D_{ur}$ to improve its capability in handling real-world scenarios. It is important to note that since $D_{ur}$ lacks reference images, we apply an unsupervised loss $\mathcal{L}_{uns}$ (Sec.~\ref{sec:exp_loss}) to guide the UIE optimization. Simultaneously, the enhanced images $\tilde{x}_r$ generated by the UIE head are used to transfer knowledge to the UOD task, enhancing object detection in clearer underwater conditions. As illustrated in Fig.~\ref{fig:overall_arch}\textcolor{blue}{(a)},  the enhanced images $\tilde{x}_r$, along with the annotated bounding boxes $b_r$ and class labels $c_r$, formulate a labeled enhanced underwater dataset $D_{le}$ that is employed to improve the UOD performance in a supervised fashion. 

% Following the Burn-In Stage, the network proceeds to the Mutual-Learning Stage. During this stage, the performances of both UIE and UOD task are further improved through mutual knowledge learning. For the UIE task, knowledge from real-world underwater images are utilized by training with unpaired real-world underwater images $D_{ur}=\{(x_r^i),i\in[1,n_r]\}$ to improve its capability in dealing with real-world situations. It's important to note that as $D_{ur}$ lacks reference images, an unsupervised loss $\mathcal{L}_{uns}$ (see in Sec.~\ref{sec:exp_loss}) is applied to guide the UIE optimization at this stage. At the same time, the enhanced images $\tilde{x}_r^i$ generated by the UIE head are utilized to transfer knowledge to the UOD task, enhancing object detection in a clearer underwater setting. As illustrated in Fig.~\ref{fig:overall_arch}\textcolor{blue}{(a)},  the enhanced images $\tilde{x}_r^i$, together with the annotated bounding boxes $b_r^i$ and class labels $c_r^i$, constitute a labeled enhanced underwater dataset $D_{le}=\{(\tilde{x}_r^i, b_r^i, c_r^i)\}$, which is utilized to train the UOD in a supervised fashion. 
% In this stage, the performance of UIE is improved by getting knowledge from real-world underwater images, and the performance of UOD is improved by getting knowledge from 

(3) \textbf{Domain-Adaptation} Stage: In this stage, our goal is to mitigate domain discrepancies and enhance UOD performance in diverse underwater environments. It is assumed that after sufficient training iterations, the domain gap in the enhanced real-world underwater images generated by the UIE head has been effectively reduced~\cite{wen2023syrea}. Consequently, the feature embeddings of the enhanced images $E(\tilde{x}_r)$ can be considered domain invariant, while the embeddings of the original real-world underwater images $E(x_r)$ are adjusted to align with $E(\tilde{x}_r)$. Instead of employing adversarial learning for feature alignment~\cite{li2022cross}, which requires additional network architectures and increases training complexity, we adopt a simpler approach by fixing the domain invariant embeddings $E(\tilde{x}_r)$ and reducing the domain shift of $E(x_r)$ with our proposed domain-adaptation loss $\mathcal{L}_{da}$ (Sec.~\ref{sec:exp_loss}).

\begin{algorithm}
    \caption{Multi-Stage Training Strategy}\label{algo}
    \begin{algorithmic}[1]
    \small
        % \REQUIRE Burin-In step $N_b$, maximum step $N_m$, paired synthetic underwater images $D_{ps} = \{(X_s, \hat{X}_s)\}$, labeled real-world underwater images $D_{ro}=\{(X_r, B_r, C_r)\}$.
        \REQUIRE Datasets $D_{ps}, D_{lr}, D_{ur}$; Burn-In step $N_b$, Mutual-Learning step $N_m$, Maximum step $N$, $N_b < N_m < N$.
        \ENSURE Final EnYOLO network $\Theta^{N}$.
        \STATE Initialize EnYOLO network. $\Theta^{0}$.
        \FOR{$i \leftarrow 0$ to  $N$}
          \STATE Feed with $D_{ps}$ and $D_{lr}$.
          \STATE Compute $\mathcal{L}_{total} \leftarrow \mathcal{L}_{enh} + \mathcal{L}_{det_r}$.
          \IF{$N_b \leq i$}
            \STATE Feed with $D_{ur}$, and formulate $D_{le}$, then feed $D_{le}$.
            \STATE Compute $\mathcal{L}_{total} \leftarrow \mathcal{L}_{total} + \mathcal{L}_{uns} + \mathcal{L}_{det_e}$.
          \ENDIF
          \IF{$N_m \leq i $}
            \STATE Get $E(\tilde{x}_r)$ and $E(x_r)$.
            \STATE Compute $\mathcal{L}_{total} \leftarrow \mathcal{L}_{total} + \mathcal{L}_{da}$.
          \ENDIF
          \STATE Optimize EnYOLO by minimizing $\mathcal{L}_{total}$.
        \ENDFOR
    \end{algorithmic}
    \vspace{-0.2em}
\end{algorithm}

\subsection{Loss Design}\label{sec:exp_loss}
As shown in Alg.~\ref{algo}, the total loss $\mathcal{L}_{total}$ is the accumulation of losses at different training stages, so we will introduce the loss functions at each training stage sequentially.

\subsubsection{Burn-In Stage}
In this stage, the UIE task is trained in a supervised manner with paired dataset $D_{ps}=\{(x_s, \hat{x}_s)\}$, then the enhancement loss $\mathcal{L}_{enh}$ is formulated as:
% In this stage, considering the supervised training of the UIE task using paired synthetic underwater images $D_{ps}=\{(x_s, \hat{x}_s)\}$, the enhancement loss $\mathcal{L}_{enh}$ is formulated as:
\begin{equation}
    \small \mathcal{L}_{enh} = ||\tilde{x}_s - \hat{x}_s||
\end{equation}
where $\tilde{x}_s$ denotes the enhanced synthetic image generated by UIE head, and $||\cdot||$ represents the L1 norm.

The UOD task is optimized with labeled real-world dataset $D_{lr}=\{(x_r, b_r, c_r)\}$. Then, the detection loss $\mathcal{L}_{det_r}$ is:
\begin{equation}
    \small \mathcal{L}_{det_r} = \mathcal{L}^{c}(x_r, b_r, c_r) + \mathcal{L}^{b}(x_r, b_r, c_r) + \mathcal{L}^{o}(x_r, b_r, c_r),
\end{equation}
where $\mathcal{L}^{c}$ is the classification loss, $\mathcal{L}^{b}$ is the bounding box loss, and $\mathcal{L}^{o}$ is the objectness loss~\cite{Jocher2020YOLOv5}.

\subsubsection{Mutual-Learning Stage}
In this stage, besides $D_{ps}$, the UIE task is also optimized with an unpaired real-world dataset $D_{ur}$. It is assumed that the enhanced real-world images should comply with certain principles inherent to clear natural images. Inspired by~\cite{jamieson2023deep}, we propose an unsupervised loss $\mathcal{L}_{uns}$ based on ``gray-world" assumption, expressed as:
\begin{equation}
    \small \mathcal{L}_{uns} = \frac{1}{3}\sum_c\left(\left(\frac{1}{N}\sum_{i, j}\tilde{x}_r(i, j)\right) - 0.5\right)^2,
\end{equation}
where $\tilde{x}_r$ represents the enhanced real-world underwater images, $c$ is the color channel, $N$ is the number of pixels, and $(i, j)$ denotes the pixel position. 

For the UOD task, we replace $x_r$ with $\tilde{x}_r$, then the detection loss $\mathcal{L}_{det_e}$ is expressed as:
\begin{equation}
    \small \mathcal{L}_{det_e} = \mathcal{L}^{c}(\tilde{x}_r, b_r, c_r) + \mathcal{L}^{b}(\tilde{x}_r, b_r, c_r) + \mathcal{L}^{o}(\tilde{x}_r, b_r, c_r),
\end{equation}

\subsubsection{Domain-Adaptation Stage}
In this stage, we align $E(x_r)$ with $E(\tilde{x}_r)$ by our proposed domain-adaptation loss $\mathcal{L}_{da}$. Inspired by~\cite{sun2016deep}, $\mathcal{L}_{da}$ reduces the domain discrepancy between $E(\tilde{x}_r)$ and $E(x_r)$ by minimizing their mean-squared error and covariance distances, which is:
\begin{equation}
    \small \mathcal{L}_{da} = \small\text{MSE}(E(x_r), E(\tilde{x}_r)) + ||C(E(x_r)) - C(E(\tilde{x}_r))||^2_{F},
\end{equation}
where $C(\cdot)$ denotes the feature covariance matrice, $\text{MSE}(\cdot)$ represents the mean-squared error, and $||\cdot||^2_F$ is the squared matrix Frobenius norm.

\section{EXPERIMENTS} \label{sec:experiments}
% In this section, we first introduce the experiment setups (Sec. \ref{sec:exp_implement_details}). Then we evaluate the UIE performance of our proposed method. Subsequently, we evaluate the UOD performance of our proposed method. Then we conduct some ablation studies to evaluate each module of our proposed method. Finally, we conduct the deployment in the onboard system for AUVs and compare its performance.

% \begin{figure*}[t]
%     % \setlength{\belowcaptionskip}{-0.5cm}
%     \centering
%     \includegraphics[width=0.95\textwidth]{fig-qualitative_comparison2.pdf}
%     \captionsetup{font={small}}
%     \caption{Qualitative comparison of the performances of \textit{\textbf{SyreaNet}} and other SOTA UIE methods on real underwater images.}
%     \label{fig6}
% \end{figure*}

\subsection{Implementation Details} \label{sec:exp_implement_details}
\noindent \textbf{Datasets}: The paired synthetic underwater dataset $D_{ps}$ employed for UIE are extracted from the Syrea dataset~\cite{wen2023syrea}. This dataset comprises a total of 20,688 training pairs. The labeled dataset used for UOD is derived from DUO dataset~\cite{liu2021dataset}, with 6,671 training images and 1,111 testing images. The DUO dataset covers four distinct labeled categories: holothurian, echinus, scallop, and starfish.

\noindent\textbf{Training Setups}: The YOLOv5 detector~\cite{Jocher2020YOLOv5} serves as our baseline network, and CSPDarkNet53~\cite{bochkovskiy2020yolov4} is employed as the backbone for our proposed EnYOLO. The network is trained using the SGD optimizer for a total of $N=150k$ steps with the Burn-In step set to $N_b=80k$ and the Mutual-Learning step set to $N_m=120k$. The base learning rate is set as $l_r=0.02$, decaying by a factor of 0.1 at the epoch $N_b$ and $N_m$, respectively. For the UOD task, a batch-size of 16 is employed, while a batch-size of 8 is utilized for the UIE task. Our framework is trained using PyTorch on two NVIDIA 3090 GPUs.

\begin{figure*}[th]
    \centering
    \includegraphics[width=0.92\textwidth]{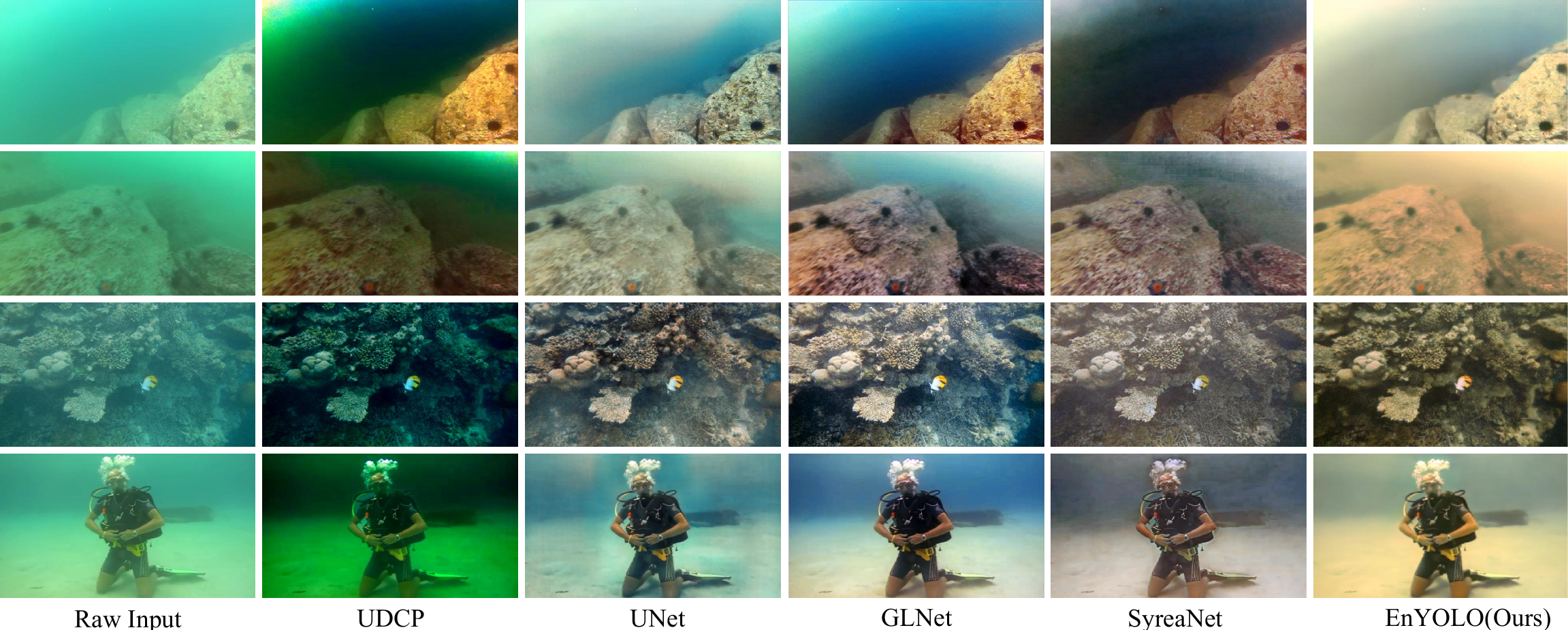}
    \caption{\small Visual comparison between various UIE methods. The first two rows contain sample images from the DUO test set~\cite{liu2021dataset}, and the last two rows contain sample images from the UIEB dataset~\cite{li2019underwater}. Zoom in for a better view.}
    \label{fig:enhancement_res}
    \vspace{-1.0em}
\end{figure*}

\subsection{Experiment on Underwater Image Enhancement}
We evaluate the UIE performance of our proposed EnYOLO through a comprehensive comparison with other SOTA UIE techniques using real-world underwater images, which are selected from the UIEB dataset~\cite{li2019underwater} and the DUO Test set~\cite{liu2021dataset}. The compared methods include the traditional approach UDCP~\cite{drews2013transmission} and learning-based methods such as UNet~\cite{falk2019u}, GLNet~\cite{fu2020underwater}, and SyreaNet~\cite{wen2023syrea}.

Fig.~\ref{fig:enhancement_res} illustrates the visual comparisons. As shown, UDCP fails to effectively address the greenish/bluish color cast, and SyreaNet introduces noticeable artifacts. While UNet and GLNet can produce visually appealing results, they introduce obvious artifacts (Row 4 for UNet) and shift the original greenish color cast to a bluish hue (Row 1 and 4 for GLNet), which could be detrimental for UOD (Sec.~\ref{sec:exp_uod}). In contrast, our proposed EnYOLO not only effectively mitigates the greenish/bluish effect but also ensures no artifacts that could potentially confound UOD task are introduced. It's worth noting that while EnYOLO demonstrates impressive performance in handling real-world underwater images across various environments, it may not excel as much in restoring fine details. This trade-off is necessary to strike a balance between the objectives of both UIE and UOD tasks. We also conduct a quantitative evaluation using Image Quality Assessment (IQA) metrics specifically designed for underwater images, including UIQM~\cite{panetta2015human}, UCIQE~\cite{yang2015underwater}, and the learning-based URanker~\cite{yang2015underwater}. The results are presented in Tab.~\ref{tab:enhance_quan}, demonstrating comparable performance by our method EnYOLO against other SOTA methods.

% A quantitative comparison is further conducted by computing some Image Quality Assessment (IQA) metrics designed for underwater images, which encompass UIQM~\cite{panetta2015human}, UCIQE~\cite{yang2015underwater}, and the learning-based metric URanker~\cite{guo2023underwater}. The results are shown in Tab.~\ref{tab:enhance_quan}, illustrating that our EnYOLO framework achieves comparable performance to other SOTA UIE methods.

\begin{table}[h]
  \centering
  \caption{\small Quantitative comparison of UIE performance for EnYOLO and other SOTA methods.}
  \label{tab:enhance_quan}
  \setlength{\tabcolsep}{3mm}{
  \begin{tabular}{c|c|c|c}%一个c表示有一列，格式为居中显示(center)
    \toprule[1.5pt]
    Methods & UIQM$\uparrow$ & UCIQE$\uparrow$  & URanker$\uparrow$\\ 
    \midrule[1.5pt]
    UDCP & 1.290 & 0.556 & 1.411 \\
    % \midrule
    UNet & 1.371 & 0.553 & 1.528 \\
    % \midrule
    GLNet & \underline{1.541} & \textbf{0.619} & \underline{1.812} \\
    % \midrule
    SyreaNet & \textbf{1.656} & \underline{0.582} & 1.781\\
    \midrule
    \textbf{EnYOLO} & 1.523 & 0.571 & \textbf{1.830} \\
    \bottomrule[1.5pt]
    \end{tabular}}
    \vspace{-0.5em}
\end{table}

\begin{table*}[t!]
  \centering
  \caption{\small Quantitative comparison of UOD performances for various methods on DUO Test set. The subscripts \textit{ho}, \textit{ec}, \textit{sc}, and \textit{st} respectively represent distinct categories: \textit{holothurian}, \textit{echinus}, \textit{scallop}, and \textit{starfish}. The subscripts \textit{green}, \textit{blue}, and \textit{turbid} denote greenish, bluish and turbid underwater environments synthesized from DUO Test set.}
  \label{tab:det_quan}
  \begin{tabular}{c|c|c|cccc|ccc}%一个c表示有一列，格式为居中显示(center)
    \toprule[1.5pt]
    Method & Backbone & mAP  & $\text{mAP}_{\textit{ho}}$ & $\text{mAP}_{\textit{ec}}$ & $\text{mAP}_{\textit{sc}}$ & $\text{mAP}_{\textit{st}}$ & $\text{mAP}_{\textit{green}}$ & $\text{mAP}_{\textit{blue}}$ & $\text{mAP}_{\textit{turbid}}$ \\ 
    \midrule[1.5pt]
    YOLOv5(baseline) & CSPDarkNet53 & 58.08 & 58.18 & \underline{64.88} & 43.45 & \underline{66.81} & 39.26 & 12.04 & 32.48 \\
    FasterRCNN  & ResNet50 & 57.41 & 59.56 & 62.12 & 41.69 & 65.38 & 25.54 & 9.67 & 26.94 \\
    % \midrule
    % \midrule
    CSAM & CSPDarkNet53 & 57.81 & \underline{60.76} & 63.61 & 42.41 & 65.20 & 38.31 & 12.23 & 32.34 \\
    JADSNet & ResNet50 & \underline{58.53} & 60.51 & 62.62 & \underline{45.70} & 65.51 & 39.41 & 13.42 &  31.85 \\
    % \midrule
    IA-YOLO & CSPDarkNet53 & 58.28 & 60.52 & 62.44 & 44.27 & 65.90 & 40.18 & 14.77 &  33.91 \\
    DENet & CSPDarkNet53 & 58.44 &  \textbf{61.17} & 63.90 & 45.55 & 65.91 & 39.67 & 15.02 & 34.07 \\
    \midrule
    GLNet+YOLOv5 & CSPDarkNet53 & 57.21 & 58.02 & 63.77 & 42.81 & 66.11 & \underline{42.31} & 16.68 & \underline{34.63} \\
    % \midrule
    SyreaNet+YOLOv5 & CSPDarkNet53 & 57.18 & 57.40 & 63.14 & 42.78 & 65.24 & 40.33 & \underline{17.81} & 34.23 \\
    % GLNet+FasterRCNN & ResNet50 & 56.94 &  58.92 & 62.41 & 41.49 & 64.36 & 30.39 & 12.07 & 31.29 \\
    % SyreaNet+FasterRCNN  & ResNet50 & 56.53 & 58.50 & 62.23 & 42.41 & 64.31 & 31.30 & 12.17 &  32.21 \\
    \midrule
    \textbf{EnYOLO}(Ours) & CSPDarkNet53 & \textbf{60.71} &  60.29 & \textbf{65.57} & \textbf{48.45} & \textbf{68.54} & \textbf{45.73} & \textbf{33.23} & \textbf{44.31} \\
    \bottomrule[1.5pt]
    \end{tabular}
    \vspace{-1.0em}
\end{table*}

\subsection{Experiment on Underwater Object Detection} \label{sec:exp_uod}
We also conduct a comprehensive UOD evaluation of our proposed EnYOLO against various SOTA approaches, including the baseline YOLOv5~\cite{Jocher2020YOLOv5}, FasterRCNN~\cite{ren2015faster}, CSAM~\cite{jiang2021underwater}, JADSNet~\cite{cheng2023joint}, IA-YOLO~\cite{liu2022image}, DENet~\cite{qin2022denet}, and \textit{Enhance}+YOLOv5. CSAM and JADSNet are two detection methods specifically designed for underwater scenarios. IA-YOLO and DENet were originally developed to address adverse challenging weather conditions in terrestrial environments. To assess their adaptability to underwater environments, we utilized their official source code and conducted retraining using the same training dataset. For \textit{Enhance}+YOLOv5, we employed SOTA UIE methods GLNet~\cite{fu2020underwater} and SyreaNet~\cite{wen2023syrea} as the pre-processing step, and the resulting enhanced images are subsequently employed to train YOLOv5 for UOD.

Tab.~\ref{tab:det_quan} presents a thorough comparison of mAP between EnYOLO and other SOTA detection techniques. Our EnYOLO achieves a significant improvement of $+2.63\%$ in mAP over the baseline network, outperforming all other SOTA detection methods. It's noteworthy that utilizing UIE methods such as GLNet or SyreaNet as a pre-processing step yields a decreased mAP score. This indicates that while the enhancement methods aim to improve visual quality, they can have an adverse impact on the detection performance. 

To further evaluate the domain adaptability of these methods, we synthesize underwater images of different environments using the DUO Test set, following the technique adopted in~\cite{wen2023syrea}. Visual comparisons are presented in Fig.~\ref{fig:demo}. The last three columns of Tab.~\ref{tab:det_quan} show that all methods experienced a performance drop when operating in new environments, and the most substantial drop occurs in the bluish environment ($-40.04\%$ for YOLOv5), mainly due to the domain discrepancies from the training set. Our proposed EnYOLO outperforms all other methods significantly across various environments, achieving an impressive $+21.19\%$ increase in the challenging bluish environment compared to the baseline, demonstrating the remarkable adaptability of our method to diverse underwater environments. 

% It's also worth mentioning that, although $\textit{Enhance}$+YOLOv5 underperforms YOLOv5 on the original test set, they exhibited slight improvements over their baseline models when dealing with new environments, which can be attributed to the enhancement methods' ability to mitigate the domain shift across different underwater conditions.

\subsection{Ablation Study}
\subsubsection{Effectiveness of UIE head} 
The effectiveness of the UIE task for UOD can be tested by removing the UIE head, which is the same with baseline model (Ab$_1$ in Tab.~\ref{tab:ablation}).

% Effectiveness of the the Domain-Adaptation.
\begin{table}[h] \scriptsize
  \centering
  \caption{\small Ablation studies on various modules.}
  \label{tab:ablation}
  \begin{tabular}{l|ccccc|c}%一个c表示有一列，格式为居中显示(center)
    \toprule[1.5pt]
    Ablations        & Ab$_1$ & Ab$_2$     & Ab$_3$     & Ab$_4$     & Ab$_5$ &EnYOLO\\
    \midrule
    UIE head         &        & \checkmark & \checkmark & \checkmark & \checkmark & \checkmark\\
    Burn-In          &        &            & \checkmark & \checkmark & \checkmark & \checkmark\\
    Mutual-Learn.  &        & \checkmark &            &    \checkmark &  & \checkmark\\
    Domain-Adapt.&        & \checkmark &            &  & \checkmark           & \checkmark\\
    \midrule
    mAP              &  58.08 &   48.21    &  57.52     & \underline{59.82}     &   59.73    & \textbf{60.71} \\
    mAP$_{green}$    &  39.26 &   25.31    &  40.13     & 42.33 &   \underline{43.25}    & \textbf{45.71} \\ 
    mAP$_{blue}$     &  12.04 &   10.29    &  25.45     & 28.97 & \underline{29.88}      & \textbf{33.23} \\
    mAP$_{turbid}$   &  32.48 &   23.86    &  39.47     &  \underline{41.25}     &  40.32     & \textbf{44.31} \\
    \bottomrule[1.5pt]
    \end{tabular}
    \vspace{-1.0em}
\end{table}

\subsubsection{Effectiveness of each training stage} 
\begin{figure}[t!]
    \centering
    \includegraphics[width=0.45\textwidth]{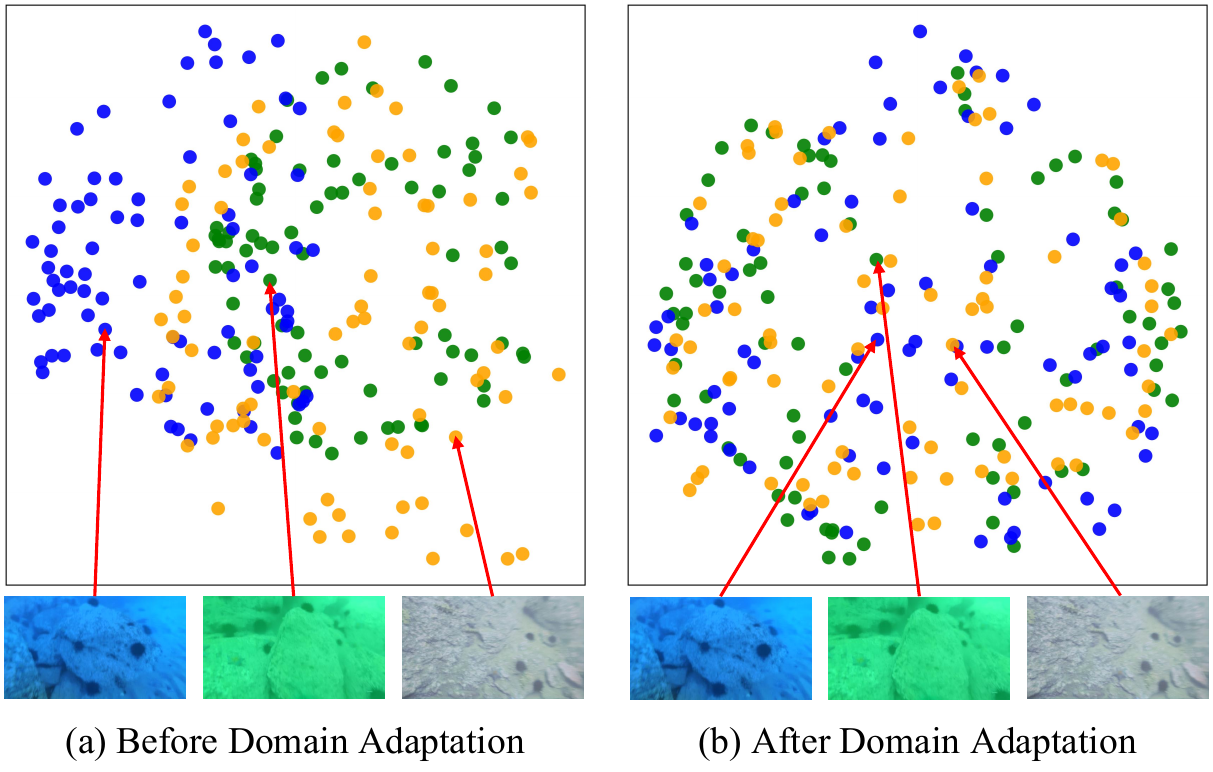}
    \caption{\small Visualization of tSNE feature embeddings from various underwater images. \text{Green}, \text{blue}, and \text{orange} dots denote features of images from greenish, bluish, and turbid underwater environments, respectively.}
    \label{fig:feat_align}
    \vspace{-1.0em}
\end{figure}

The effectiveness of the Burn-In stage can be seen in Ab$_2$ of Tab.~\ref{tab:ablation}, which shows a significant decrease in mAP of $-9.87\%$ to Ab$_1$, indicating the requirement for the network to acquire basic abilities before further refinement. In contrast, Ab$_3$ exhibits only a marginal decrease in mAP compared to Ab$_1$. The effectiveness of Mutual-Learning and Domain-Adaptation stages can be verified in the results of Ab$_4$ and Ab$_5$ in Tab.~\ref{tab:ablation}, respectively. We further visualize the influence of domain-adaptation by visualizing the backbone feature embeddings of different underwater images. As shown in Fig.~\ref{fig:feat_align}, the feature embeddings from different underwater environments show a mixed pattern after domain-adaptation.

\subsection{Framework Flexibility and Efficiency Analysis} \label{sec:exp_eff}
We introduce three modes for real-world AUV testing: Detection, Enhance, and Dual Modes. These options allow for more flexible operation. In Detection/Enhance Mode, the AUV can only employ a single task based on specific requirements, significantly reducing computational cost. In Dual Mode, both UIE and UOD tasks run simultaneously. This enables researchers to visualize detections alongside enhanced underwater images, thus facilitating a more comprehensive understanding of the detections.

% To better fit the real-world application for AUVs, our framework could be conducted in three different modes, including Detection Mode, Enhance Mode, and Dual Mode. As the UIE and UOD heads could be decoupled during testing, then in either Detection or Enhance Mode, the AUV could runs solely with single task head, which could largely save the onboard computational cost and reduce the inference latency. In Dual Mode, researchers could visualize the detection result in enhanced underwater images, which could help researchers better understand the result.

We then conduct an efficiency analysis. As illustrated in Tab.~\ref{tab:latency}, in the Dual Mode, the inference latency is only slightly increased by $3.17$ms compared to the Detection Mode, achieving an impressive frame rate of $74.29$ fps. It significantly outperforms all other methods for simultaneous enhancement and detection (Row 3 to 5), highlighting our proposed method's superior suitability for deployment on AUV's onboard systems.

% . The results highlight the superior suitability of our proposed method for deployment in the onboard systems of AUVs.

\begin{table}[h]\scriptsize
  \centering
  \caption{\small Efficiency Analysis of our framework and other methods (input size $640 \times 640$, on a single NVIDIA RTX-3090 GPU)}
  \label{tab:latency}
  \begin{tabular}{c|c|c|c|c}%一个c表示有一列，格式为居中显示(center)
    \toprule[1.5pt]
    Methods    & Latency $\downarrow$ & FPS $\uparrow$ & GFLOPs $\downarrow$ & Params $\downarrow$ \\
    \midrule[1.5pt]
    YOLOv5     & \textbf{10.29} ms  & \textbf{97.15} &  \textbf{20.04} & \textbf{20.88} M \\
    FasterRCNN & 20.44 ms & 48.92 &  91.30 & 41.75 M \\
    IA-YOLO    & 64.41 ms & 15.53 &  192.81& 61.70 M \\
GLNet+YOLOv5   & 93.09 ms & 10.74 &  75.39 & 66.18 M \\
SyreaNet+YOLOv5& 74.87 ms & 13.36 &  112.53& 49.88 M \\
    \midrule
EnYOLO (Det. Mod.) &\textbf{10.29} ms&  \textbf{97.15} & \textbf{20.04}  & \textbf{20.88} M  \\
EnYOLO (Enh. Mod.) &\underline{11.71} ms&  \underline{85.40} & \underline{21.85}  & \underline{22.85} M  \\
EnYOLO (Dual Mod.) &13.46 ms&  74.29 &  24.04 &  31.57 M \\
    \bottomrule[1.5pt]
    \end{tabular}
    \vspace{-1.0em}
\end{table}

\section{CONCLUSIONS} \label{sec:conclusions}

In this study, we propose EnYOLO, a unified real-time framework designed for simultaneous UIE and UOD with domain-adaptive capability. The UIE and UOD heads share the same backbone and employ lightweight designs. To balance dual training, we introduce a multi-stage training strategy aimed at consistently improving the performance of both tasks. A novel domain-adaptation approach to mitigate domain shifts for UOD is also proposed. Extensive experimentation attests that EnYOLO achieves SOTA performance in both UIE and UOD tasks, while demonstrating superior UOD adaptability across varying underwater environments. The efficiency analysis highlights EnYOLO's impressive real-time performance, demonstrating its substantial potential for onboard deployment.

\bibliographystyle{ieeetr}
% \biboptions{numbers,sort&compress}
\bibliography{IEEEexample}

\end{document}